\documentclass[11pt,a4paper]{article}
\usepackage[hyperref]{acl2021}
\usepackage{times}
\usepackage{latexsym}

\usepackage{url}
\usepackage{hyperref}
\usepackage{inputenc}
\usepackage{caption}
\usepackage{graphicx}
\usepackage{amsmath}
\usepackage{booktabs}
\usepackage{subcaption}
\usepackage{multirow}

\usepackage{hyperref}

\makeatletter
\newcommand{\printfnsymbol}[1]{%
  \textsuperscript{\@fnsymbol{#1}}%
}
\makeatother

\usepackage{microtype}

\aclfinalcopy 

\title{LOA: Logical Optimal Actions for Text-based Interaction Games}

\author{Daiki Kimura\thanks{\:\: denotes equal contribution} \:\: Subhajit Chaudhury\printfnsymbol{1} \:\: Masaki Ono \:\: Michiaki Tatsubori \\
{\bf Don Joven Agravante} \:\: {\bf Asim Munawar} \:\: {\bf Akifumi Wachi} \:\: {\bf Ryosuke Kohita} \:\: {\bf Alexander Gray} \\
IBM Research\\
{\small \texttt{\{daiki, subhajit, moono, mich\}@jp.ibm.com,}}\\
{\small \texttt{\{don.joven.r.agravante, asim, akifumi.wachi, kohi, alexander.gray\}@ibm.com}}
}

\date{}

\begin{document}
\maketitle
\begin{abstract}
We present Logical Optimal Actions (LOA), an action decision architecture of reinforcement learning applications with a neuro-symbolic framework which is a combination of neural network and symbolic knowledge acquisition approach for natural language interaction games. The demonstration for LOA experiments consists of a web-based interactive platform for text-based games and visualization for acquired knowledge for improving interpretability for trained rules. This demonstration also provides a comparison module with other neuro-symbolic approaches as well as non-symbolic state-of-the-art agent models on the same text-based games. Our LOA also provides open-sourced implementation in Python for the reinforcement learning environment to facilitate an experiment for studying neuro-symbolic agents. Demo site: https://ibm.biz/acl21-loa, Code: https://github.com/ibm/loa
\end{abstract}

\section{Introduction}

Neuro-symbolic~(NS) hybrid approaches have been proposed for overcoming the weakness of deep reinforcement learning~\cite{Dong+:ICLR2019:NLM,Jhang+Luo:ICML2019:NLRL,kimura2018daqn,marioirl}, including less training data with generalization, external knowledge utilization, and direct explainability of what is learned. Study of reinforcement learning~(RL) in non-symbolic environments, such as those with natural language and visionary observations, would be an important step towards the real-world application of the approaches beyond classic and symbolic environments.

Under certain controls necessary for studying RL, text-based games provide complex, interactive, and a variety of simulated environments where the environmental game state observation is obtained through the text description, and the agent is expected to make progress by entering text commands. In addition to language understanding~\cite{Ammanabrolu+Riedl:NAACL2019:KG-DQN,Adhikari+:NeurIPS2020:GATA}, successful play requires skills such as long-term memory~\cite{Narasimhan+:EMNLP2015:LSTM-DQN}, exploration~\cite{Yuan+:2018:LSTM-DRQN}, observation pruning~\cite{Chaudhury+:EMNLP2020:CREST}, and common sense reasoning~\cite{Murugesan+:AAAI2021:TWC}. However, these studies are not using the neuro-symbolic approach which is a combination of the neural network and the symbolic framework.

\begin{figure}[t]
    \centering
    \includegraphics[width=1.0\linewidth]{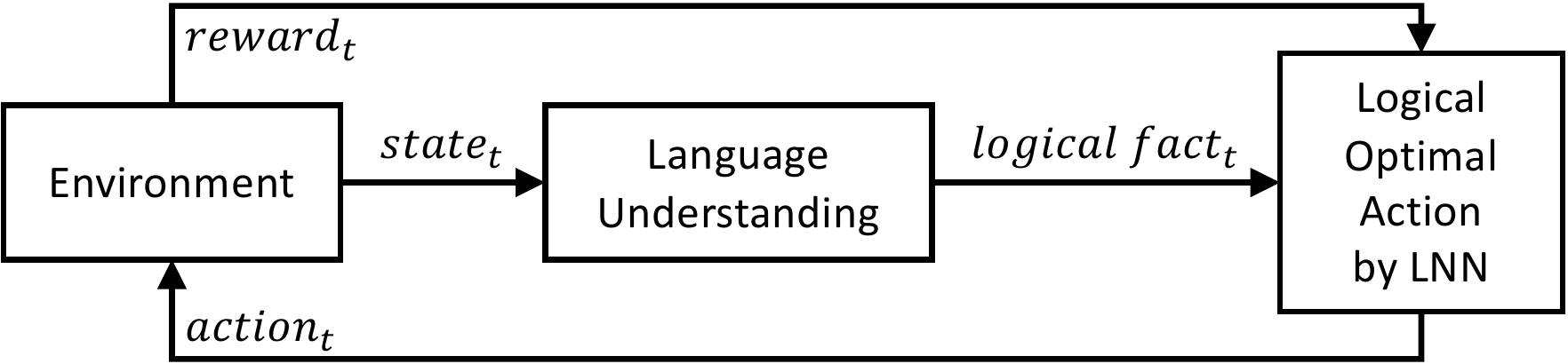}
    \caption{An architecture overview for LOA.}
    \label{fig:architecture}
\end{figure}

A recent neuro-symbolic framework called the Logical Neural Networks~(LNN)~\cite{Riegel+:2020:LNN} simultaneously provides key properties of both neural networks~(learning) and symbolic logic~(reasoning). The LNN can train the constraints and rules with logical functions in the neural networks, and since every neuron in the network has a component for a formula of weighted real-valued logics, it can calculate the probability and contradiction loss for each of the propositions. At the same time, trained LNN follow symbolic rules, which means they yield a highly interpretable disentangled representation. Using this benefit of LNN, we proposed a neuro-symbolic RL method that uses pre-defined external knowledge in logical networks, and the method successfully plays on the text-based games~\cite{Kimura+:IJCAIw2020}.

In this demonstration~(demo site: https://ibm.biz/acl21-loa), we present a Logical Optimal Actions~(LOA) architecture for neuro-symbolic RL applications with LNN~\cite{Riegel+:2020:LNN} for text-based interaction games. While natural language-based interactive agents are the ambitious but attractive target as real-world applications of neuro-symbolic, it is not easy to provide an environment for the agent. The proposed demonstration uses text-based games learning environment, called TextWorld~\cite{Cote+:CGW2018:TextWorld}, as a miniature of a natural language-based interactive environment. The demonstration provides a web-based user interface for visualizing the game interaction, which is including displaying the natural text observation from the environment, typing the action sentence, and showing the reward value from the taken action. The LOA in this demonstration also visualizes trained and pre-defined logical rules in LNN via the same interface, and this will help the human user understand the benefits of introducing the logical rules via neuro-symbolic frameworks. We also supply an open-sourced implementation for demo environment and some RL methods. This implementation contains our logical approaches and other state-of-the-art agents.

\section{Logical Optimal Action}

Our proposing LOA is an RL framework which is combining logical reasoning and neural network training. These training and reasoning are provided from functionalities of LNN~\cite{Riegel+:2020:LNN} that is simultaneously providing key properties of both neural networks and symbolic logic. Figure~\ref{fig:architecture} shows the overview architecture for LOA. The LOA model receives the logical state value as logical fact from the language understanding component which receives raw natural language state value from the environment. The model forwards into LNN for the input to get the optimal action for it, the action goes into the environment to execute the action command, then reward is input to LOA agent. The LOA will be trained the action decision network in LNN by using the acquired reward value and chosen action from the network.

\section{LOA Demo}

\begin{figure}[t]
    \centering
    \includegraphics[width=1.0\linewidth]{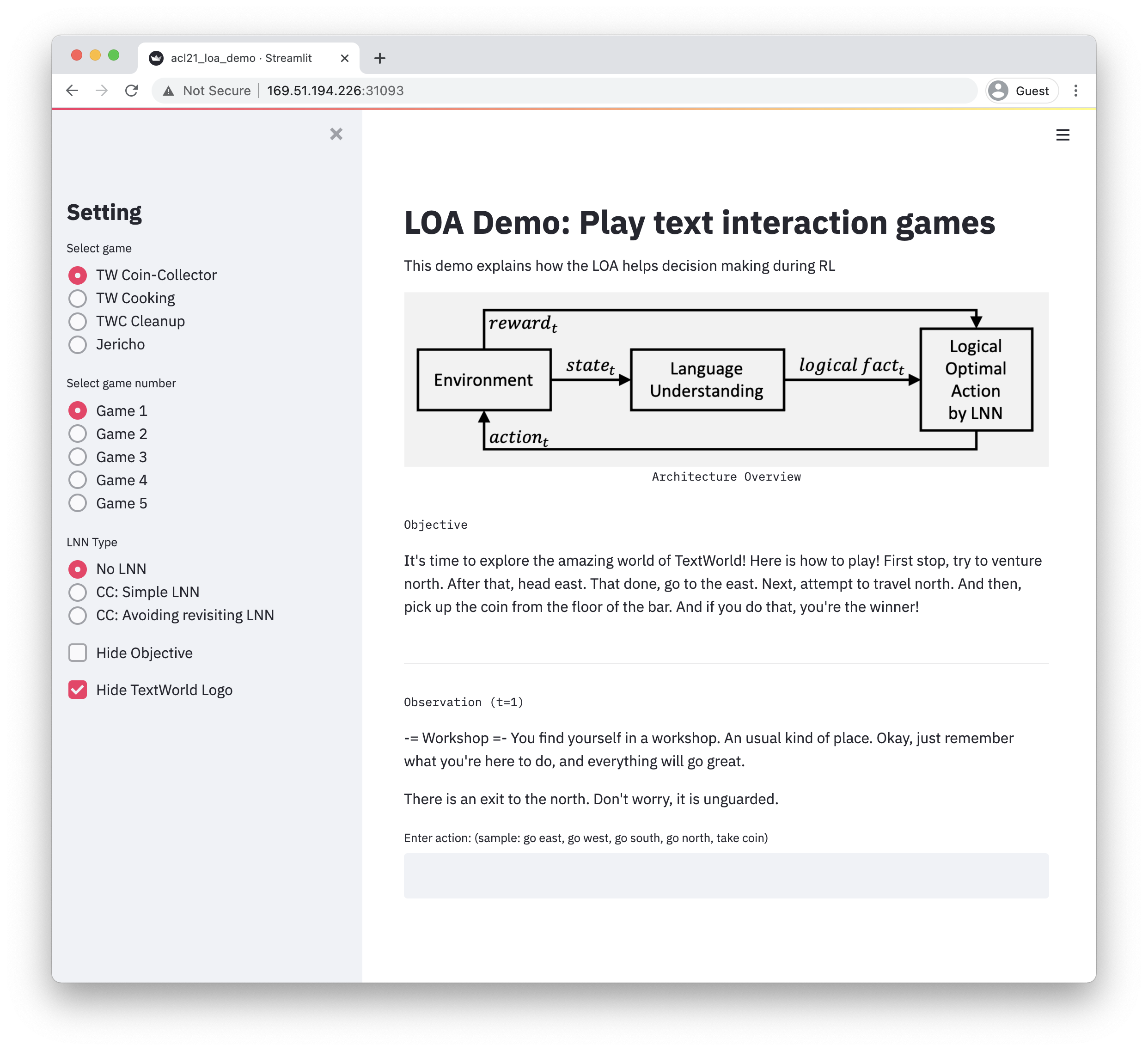}
    \caption{Initial view for LOA demo.}
    \label{fig:demo1}
\end{figure}

\begin{figure}[t]
    \centering
    \includegraphics[width=1.0\linewidth]{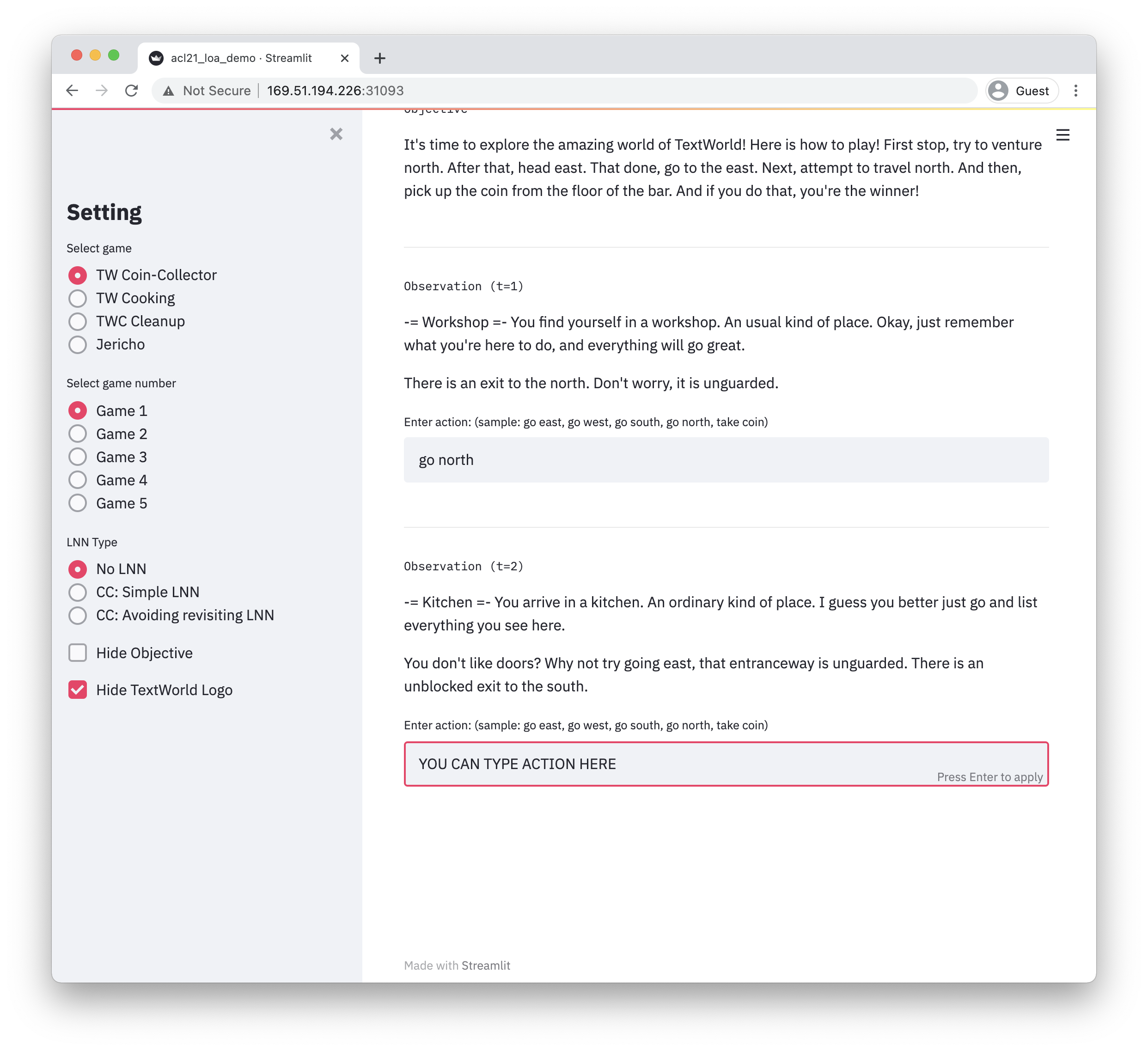}
    \caption{View for playing the game.}
    \label{fig:demo2}
\end{figure}

\begin{figure}[t]
    \centering
    \includegraphics[width=1.0\linewidth]{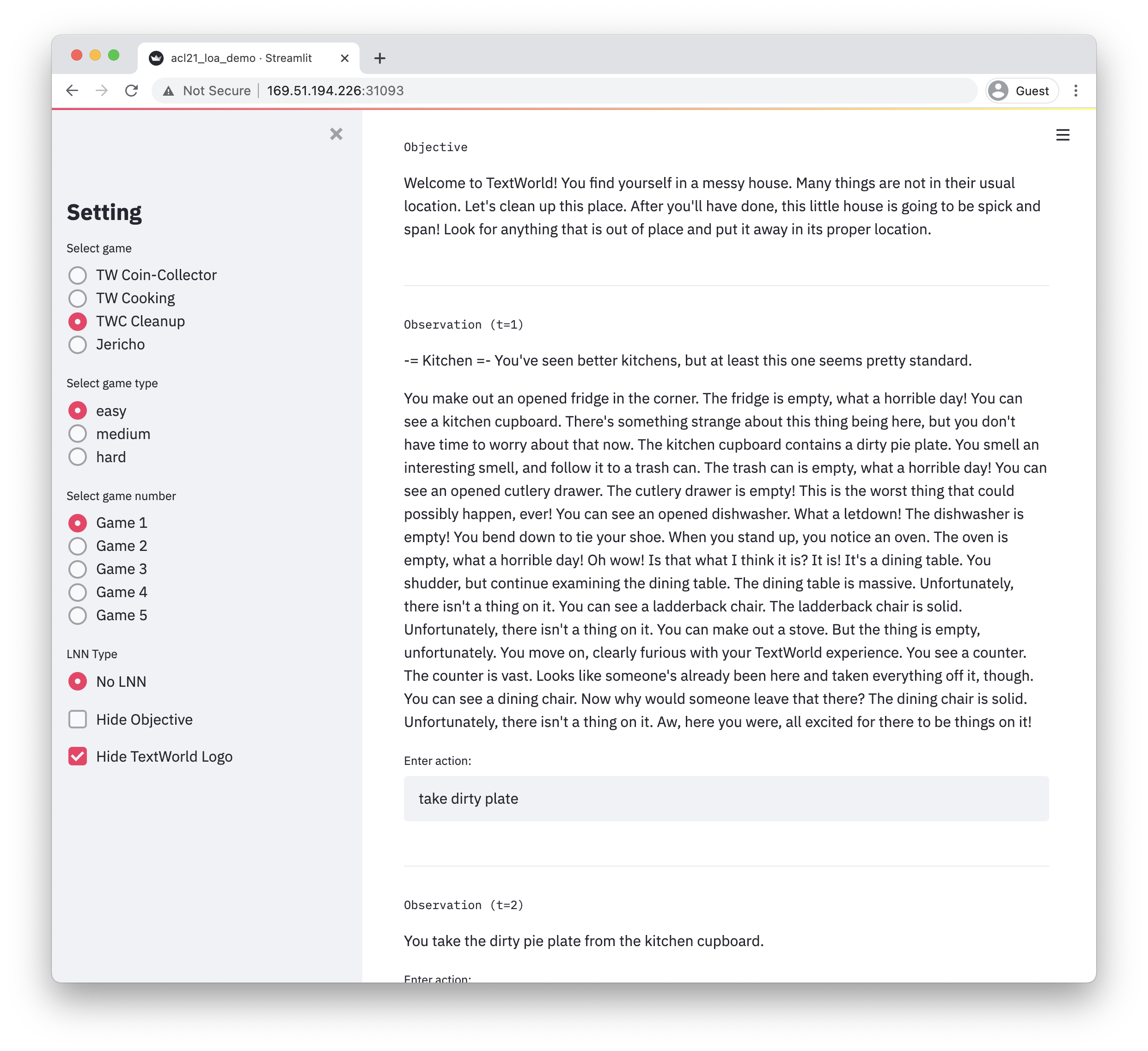}
    \caption{View for playing the cleanup game.}
    \label{fig:demo3}
\end{figure}

The proposing web-based LOA demonstration supports two functionalities: 1) play the text-based game by human interactions, 2) visualize the trained and pre-defined LNN to increase interpretability for acquired rules. 

For playing the games by web interface, Fig.~\ref{fig:demo1} shows an initial view for the LOA demonstration. On the left-hand side, we can choose the game from some existing text-based interaction games~\footnote{We are planning to add other games.}, such as TextWorld Coin-Collector game~\cite{Cote+:CGW2018:TextWorld}, TextWorld Cooking game~\cite{Cote+:CGW2018:TextWorld}, TextWorld Commonsense Cleanup game~\cite{Murugesan+:AAAI2021:TWC}, and Jericho game~\cite{hausknecht19}. Figure~\ref{fig:demo2} shows the view for playing the TextWorld game, and Fig.~\ref{fig:demo3} shows the view for another game~(cleanup task). The human player can input any action by natural language then the demonstration system displays the raw observation output from the environment. 

\begin{figure}[t]
    \centering
    \includegraphics[width=1.0\linewidth]{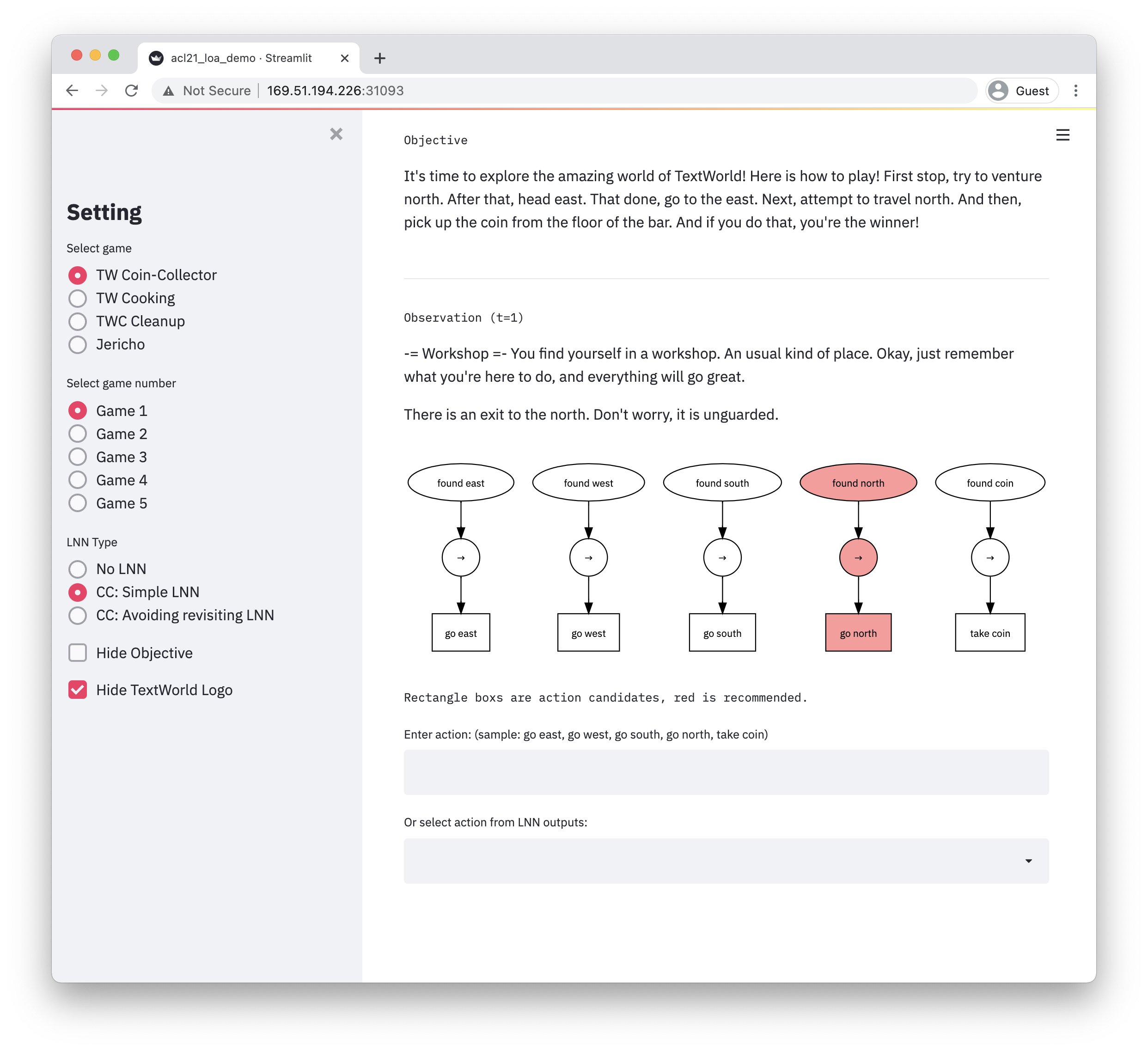}
    \caption{Displaying the simple LNN with given state.}
    \label{fig:demo4}
	\vspace{15pt}
    \includegraphics[width=1.0\linewidth]{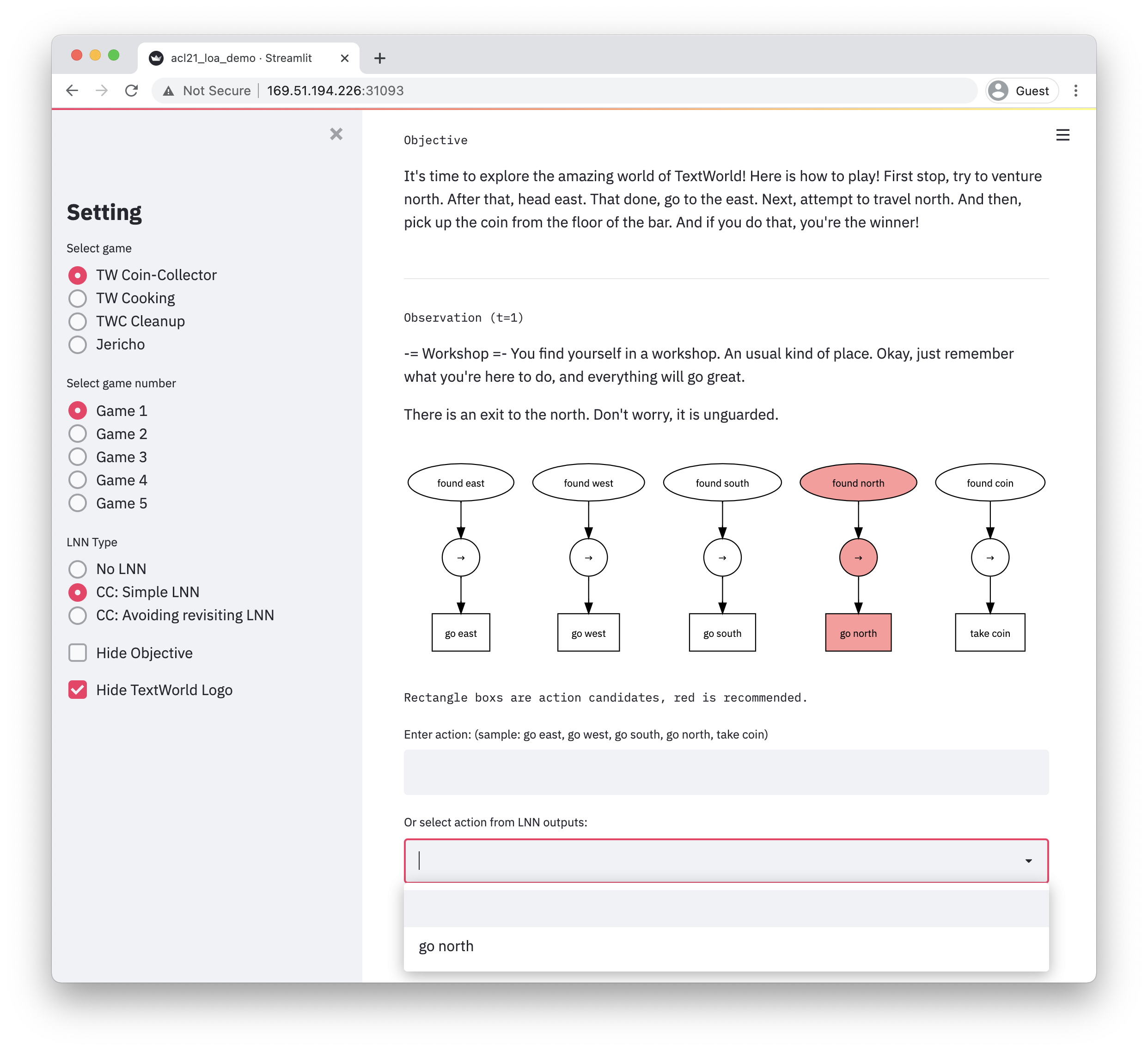}
    \caption{User can choose the recommended action.}
    \label{fig:demo5}
\end{figure}

For visualizing the trained and pre-defined neuro-symbolic network in LNN, Fig.~\ref{fig:demo4} and Fig.~\ref{fig:demo5} show the example of the LNN output. In these figures, the LNN contains simple rules for the TextWorld Coin-Collector game; for example, the rule is the agent takes `go east' action, when the agent finds the east room (``\texttt{found west}'' $\rightarrow$ ``\texttt{go west}''). The round box explains the proposition from the given observation inputs, the circle with a logical function means a logical function node of LNN, and the rectangle box explains an action candidate for the agent. The highlighted nodes~(red node) have `true' value, and non-highlighted nodes~(white node) have `false' value. In Fig.~\ref{fig:demo4}, the agent found the north exit from the given observation~(``\texttt{Observation (t=1)}'') by using semantic parser~\footnote{This parser is out of our current research topic, we prepare a simple semantic parser.}, then the going north room action~(``\texttt{go north}'') are activated. In Fig.~\ref{fig:demo5}, if the user clicks the selectable box, the LOA recommends only one action which is `go north'. In this demonstration, we show the benefit of introducing the LNN into an RL agent, we don't prepare to automatically choose the action by LOA framework. However, if we execute the RL with LOA framework, the RL agent can converge faster than other non-symbolic and neuro-symbolic methods.

\begin{figure}[t]
    \centering
    \includegraphics[width=1.0\linewidth]{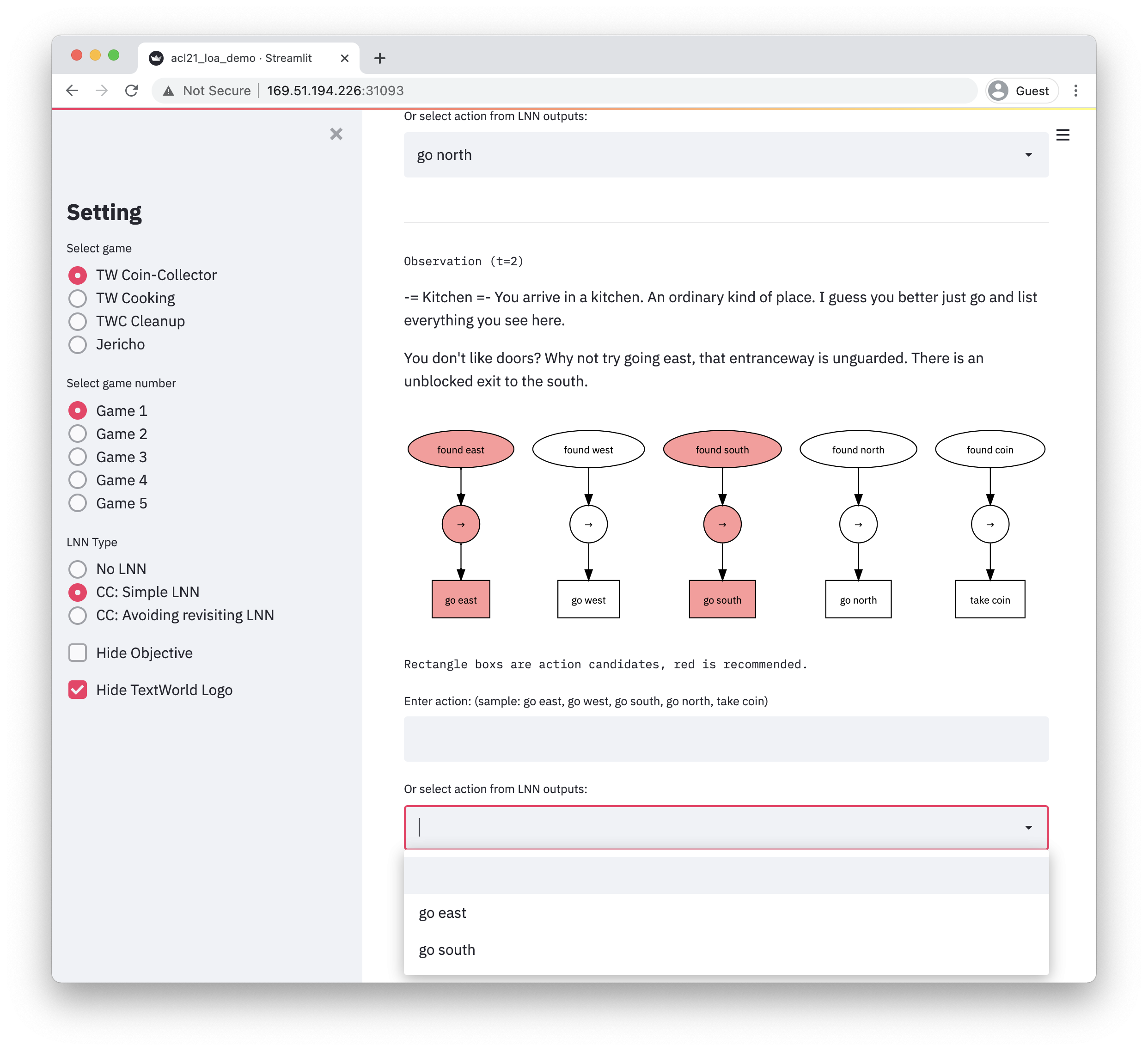}
    \caption{Result for simple LNN.}
    \label{fig:demo6}
	\vspace{15pt}
    \includegraphics[width=1.0\linewidth]{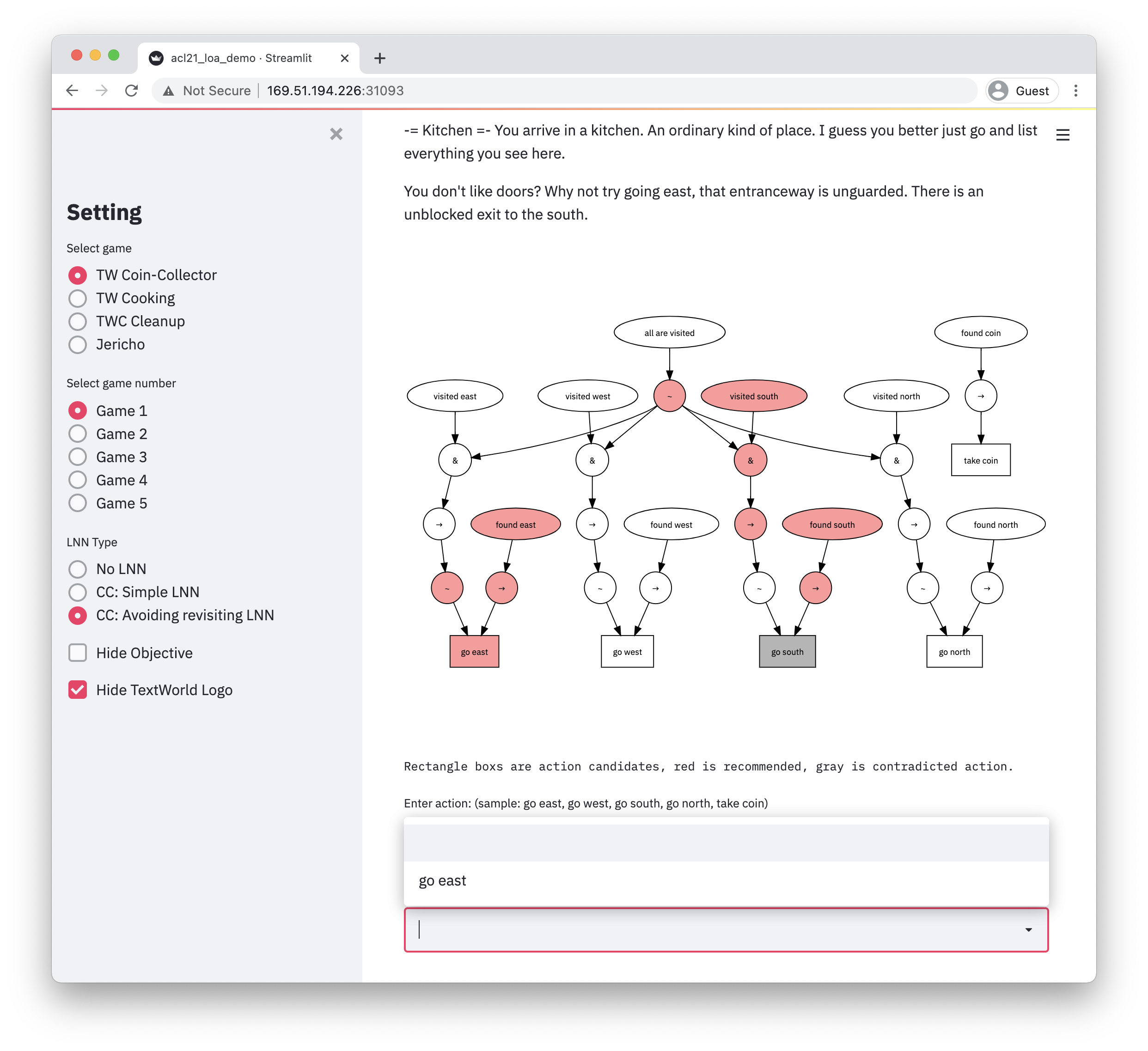}
    \caption{Result for avoiding revisiting LNN.}
    \label{fig:demo7}
\end{figure}

After selecting ``\texttt{go north}'' action at $t=1$, next observation sentence and LNN output for next step are shown in Fig~\ref{fig:demo6}. In this step, the agent found two doors, which are east and south; however, the south door is connected to the previous room because the agent took going north action at the previous step. Since this LNN is simple LNN, the ``\texttt{go south}'' action is also recommended in Fig~\ref{fig:demo6}. Figure~\ref{fig:demo7} shows the output of the complicated LNN which has functionality for avoiding revisiting the visited room. By using such the LNN, LOA can output only ``\texttt{go east}'' action by having contradiction loss in LNN. This is a benefit of introducing the neuro-symbolic framework, and the human user can easily understand the reason for the taken action by the agent with this interpretability by LOA.

\section{Conclusion}

We propose a novel demonstration~(URL: https://ibm.biz/acl21-loa) which provides to play the text-based games on the web interface and visualize the benefit of the neuro-symbolic algorithm. This application helps the human user understand the trained network and the reason for taken action by the agent. We also extend more complicated LNN for other difficult games on the demo site. At the same time, we open the source code for the demonstration~(URL: https://github.com/ibm/loa).

\bibliographystyle{acl_natbib}
\bibliography{acl2021_cameraready}

\end{document}